\documentclass[letterpaper, 10 pt, conference]{ieeeconf}  % Comment this line out if you need a4paper

\IEEEoverridecommandlockouts                              % This command is only needed if 
\usepackage{graphicx} %use graph format
\usepackage{epstopdf}
\usepackage{amsmath}
\usepackage{subfig}
\usepackage{multirow}
\usepackage{threeparttable}
\usepackage{caption}
\usepackage{blindtext}
\usepackage{amsfonts}
\usepackage{hyperref}
\usepackage{tabulary}
\newcolumntype{K}[1]{>{\centering\arraybackslash}p{#1}}

\title{\LARGE \bf
Pose Graph Optimization for Unsupervised Monocular Visual Odometry
}

\author{Yang Li$^{1}$, Yoshitaka Ushiku$^{1}$ and Tatsuya Harada$^{1,2}$ % <-this % stops a space
\thanks{ $^{1}$The authors are with the Department of Mechano-Informatics, Graduate School of Information Science and Technology, The University of Tokyo, 7-3-1 Hongo Bunkyo-ku, Tokyo, Japan 
	{\tt\small \{liyang, ushiku, harada\}@mi.t.u-tokyo.ac.jp}}%
\thanks{$^{2}$Tatsuya Harada is with RIKEN}%
}

\begin{document}

\maketitle
\thispagestyle{empty}
\pagestyle{empty}

%%%%%%%%%%%%%%%%%%%%%%%%%%%%%%%%%%%%%%%%%%%%%%%%%%%%%%%%%%%%%%%%%%%%%%%%%%%%%%%%
\begin{abstract}

Unsupervised Learning based monocular visual odometry (VO) has lately drawn significant attention for its potential in label-free leaning ability and robustness to camera parameters and environmental variations. However, partially due to the lack of drift correction technique, these methods are still by far less accurate than geometric approaches for large-scale odometry estimation. In this paper, we propose to leverage graph optimization and loop closure detection to overcome limitations of unsupervised learning based monocular visual odometry. To this end, we propose a hybrid VO system which combines an unsupervised monocular VO called NeuralBundler with a pose graph optimization back-end. NeuralBundler is a neural network architecture that uses temporal and spatial photometric loss as main supervision and generates a windowed pose graph consists of multi-view 6DoF constraints. We propose a novel pose cycle consistency loss to relieve the tensions in the windowed pose graph, leading to improved performance and robustness. In the back-end, a global pose graph is built from local and loop 6DoF constraints estimated by NeuralBundler, and is optimized over $\mathrm{SE(3)}$. Empirical evaluation on the KITTI odometry dataset demonstrates that 1) NeuralBundler achieves state-of-the-art performance on unsupervised monocular VO estimation, and 2) our whole approach can achieve efficient loop closing and show favorable overall translational accuracy compared to established monocular SLAM systems.

\end{abstract}

%%%%%%%%%%%%%%%%%%%%%%%%%%%%%%%%%%%%%%%%%%%%%%%%%%%%%%%%%%%%%%%%%%%%%%%%%%%%%%%%
\section{Introduction}
 
Nowadays, monocular visual odometry (VO) can be newly divided into two groups based on the technique and the framework adopted: geometry-based and learning-based approaches. The geometric approach is usually solved via shallow feature or photometric re-projection followed by on-line error minimization. Learning base visual odometry is a newly emerged solution and has already achieved promising results on some benchmarks. This approach is solved by off-line training of End-to-End deep neural networks driven by a large number of image sequences. Benefiting from the nature of data-driven approach and the potential of deep neural networks, learning based monocular VO has shown clear advantages over geometric methods in the following aspects: 1) no need for parameter tuning effort, 2) robustness to tracking failure and scale drift \cite{sfmleaner}, 3) capable of recovering metric scale from monocular image by using stereo image pairs in training phase \cite{undeepvo}\cite{monodepth}\cite{dvso}, 4) high potential in directly integrating semantic information for robust camera tracking \cite{why-slam-matters}\cite{fcn}.
%\end{itemize}

\begin{figure}[t]
	\includegraphics[width=0.47\textwidth]{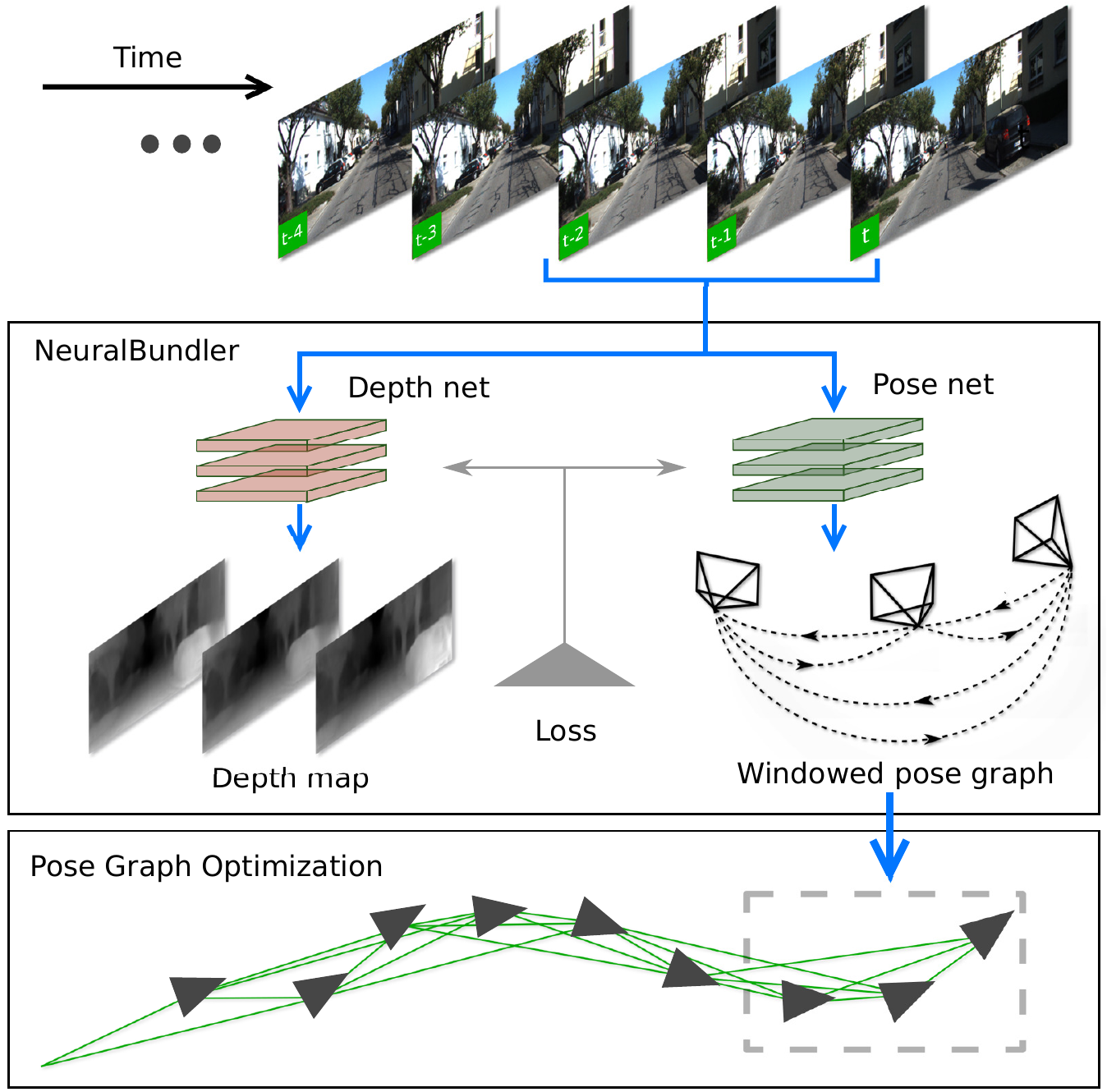}
	\caption{ Overview of the proposed VO system. 
	}
\end{figure}

Due to the existence of outliers, noises, etc., all frame-frame VO systems suffer from drifts (the accumulation of small errors over time). In the geometric lineup, the so-called graph-based SLAM mitigates this problem by combining the geometric VO with an optimization back-end to continuously regulate the landmark's positions and the camera's poses, as known as the graph optimization technique. Graph-based SLAM has achieved success in many cases with established systems, such as feature-based  PTAM \cite{ptam} and ORB-SLAM \cite{orbslam}, and direct LSD-SLAM \cite{lsdslam} and DSO \cite{dso}. Especially, with the help of loop closing, i.e. graph optimization with correctly established loop closure constraints, SLAM is able to significantly reduce global trajectory drift.

Partially due to the lack of drift correction technique, learning based VO is still by far less accurate than the geometric approach. Therefore, it is worth exploiting the potential of SLAM's graph optimization technique for learning based visual odometry. However, such work has never been done.
We presume that this is because most of the existing deep learning architectures are either trying to mimic the graph optimization process through memory-based LSTM network \cite{deepvo} or trying to integrate loop closing into a whole end-to-end learning process \cite{nerual-opt}\cite{navi-complex-envirment}. These works ignore the fact that loop closing is randomly occurred event and requires simultaneously processing of sequence with arbitrary length, which the neural networks are inadequate to do.

In this work, we build on the main idea of the unsupervised VO architecture: SfMLeaner \cite{sfmleaner}, the bag of visual word based place recognition tool: DBoW2 \cite{dbow2},
and the insight of using Covisibility / Essential Graph optimization \cite{ptam}\cite{orbslam}\cite{covi-graph}\cite{covi-graph2} for large-scale loop closing operation,  to design a hybrid VO system with the following contributions: 

\begin{itemize}
	\item An unsupervised learning based monocular visual odometry called NeuralBundler which produces a windowed pose graph from monocular image sequence with a novel training loss that enforces pose cycle consistency.
	\item Efficient loop closing procedure based on the optimization of a pose graph which is built from local and loop 6DoF constraints estimated by the proposed unsupervised monocular VO.
\end{itemize}

We present the evaluation on KITTI odometry dataset \cite{kitti}. NeuralBundler achieves the state of the art performance on learning based visual odometry estimation and our whole approach is able to perform efficient loop closing and yields favorable overall translational accuracy compared to established monocular SLAM systems. To the best of our knowledge, this is the first attempt to combine deep learning based VO with the classic graph optimization technique. Our research provides insights into the design of future SLAM system which could directly integrate the robustness and perception ability of deep neural network.

\section{Related Work}

\subsection{Geometry based Visual Odometry}
Geometry based Visual Odometry is a well-studied problem with two main solutions: feature-based and direct methods. Feature-based approach usually consists of two stages: 1) data association, through hand-engineered feature extraction (e.g. SURF \cite{surf-feature}, SIFT \cite{sift-feature} and ORB \cite{orb-feature}) and matching, and 2) pose estimation via minimizing feature point re-projection error. Direct VO treats data association and pose estimation as a whole optimization problem and solve it by minimizing the photometric error. Although these methods are effective in many cases, they are usually hard-coded and require extensive parameter tuning effort in order to ensure performance in a given scenario. The reliance on accurate data association can lead to tracking failure in regions of low texture, illumination change, and occlusions. Moreover, geometric approaches inherently suffer from the fact that camera motion can only be estimated up to an unknown scale which also leads to scale drift over time.

\subsection{Learning based Visual Odometry}

This line of research can be further divided into 2 categories: supervised and unsupervised methods.
In supervised approaches, Wang et al. \cite{deepvo} train a deep recurrent network end-to-end to predict ego-motion using ground truth trajectory as supervision. Kendall et al. \cite{pose-net} directly regress the camera's world pose from RGB images with the convolutional neural network. Zhou et al. \cite{DeepTAM} proposed a coarse to fine deep learning framework to	track camera based on key-frame.  However, the cost of collecting ground truth poses limits the application of such methods.

On the other hand, the unsupervised approach attracts more attention for its label-free leaning ability. The first architecture that achieved unsupervised learning of ego-motion from the video is SfMLeaner proposed by Zhou et al \cite{sfmleaner}. SfMLeaner takes consecutive temporal images to predict both depth and ego-motion with view synthesis as supervision. However, similar to geometric approaches, SfMLeaner can only observe ego-motion in a relative scale from monocular image. Godard et al. \cite{monodepth} show that the solution to recovery metric scale for depth prediction is using stereo images constraints for network training. Soon after, Nan et al. \cite{dvso} integrate such a monocular deep depth predictor into DSO \cite{dso} as direct virtual stereo measurements. The follow-up works from Ruihao et al. \cite{undeepvo} and Huangying et al. \cite{vo-feate} also show that leveraging stereo image pairs with known baseline in training phase enable the networks to recover metric scale for both depth and pose estimation.

Existing learning based methods only focus on frame-frame VO estimation. We propose to utilize the multi-view discrepancies within a temporal window. Our method produces a windowed pose graph and uses a novel loss to ensure pose consistency in the graph. It will be described in Section \hyperref[cycle-loss]{III-B}.

\begin{figure*}[t]
	\centering
	\includegraphics[width=1.0\textwidth]{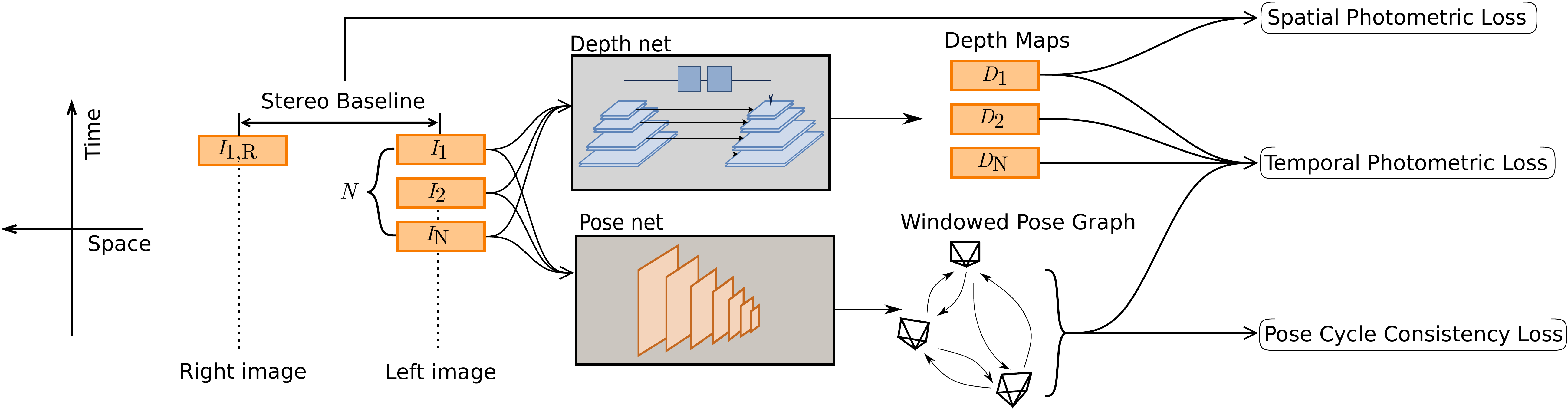}
	\caption{Training procedure of NeuralBundler. $1$ right image from the first frame and $N$ left images are involved in the training phase. Both pose net and depth net only require monocular sequence in the testing phase. }
	\label{figurelabel}
	
\end{figure*}
\subsection{Graph-based SLAM}

Graph-based SLAM maintains a global graph whose nodes represent camera's poses or landmarks and an edge represents a sensor measurement that constrains the connected poses \cite{opt-tutorial}.  
Apparently, such constraints can be conflicting to each other since measurements are easily influenced by noise.
Once such a graph is constructed, SLAM uses graph optimization method (i.e. nonlinear least-squares error minimization via the Gauss-Newton or Levenberg-Marquardt algorithm) to find a configuration of the nodes that is maximally consistent with all the constraints. The graph optimization procedure, with the presence of both camera pose and landmarks in the graph, is called Bundle-Adjustment (BA). Monocular SLAM systems that apply graph optimization include PTAM \cite{ptam}, LSD-SLAM \cite{lsdslam}, ORB-SLAM \cite{orbslam}, DSO \cite{dso}, etc.

In monocular SLAM, loop closure is solved through a pose graph optimization with 7DoF similarity constraints $(\mathrm{Sim(3)})$ to correct the scale drift \cite{sim3}.  A pose graph is built on selected key-frame connected by the pose-pose constraints. Pose-pose constraints are defined by covisiblity \cite{covisibility} and estimated by a geometry base VO front end. Two poses are connected to each other if they share enough common features. The graph built on covisiblity is called Covisiblity Graph. In order to achieve scalable, real-time performance, Raúl et al. \cite{orbslam} proposed to perform loop closing on a much lighter Essential Graph which retains all the nodes (key-frames) from Covisibility Graph and a subset of edges with high covisibility.  As shown in \cite{orbslam}, the optimization of a properly constructed Essential Graph is already very accurate that full Bundle Adjustment only makes marginal improvement. We take the idea of loop closing with graph optimization and apply it to an unsupervised learning based visual odometry. Details are shown in Section \hyperref[back-end-A]{IV}. Different from the monocular SLAM approach, we only optimize the pose graph with 6DoF constraints, i.e. $\mathrm{SE(3)}$. The reason is that perhaps owing to the nature of data-driven method, in our learning based VO, scale drift is so small, that, in the experiments, optimization over $\mathrm{SE(3)}$ and $\mathrm{Sim(3)}$ almost leads to the identical result. 

Loop closure is triggered by the place recognition technique. 
Appearance or image-image matching based methods, such as the bag of word approaches FAB-MAP \cite{fab-map} and DBoW2 \cite{dbow2}, are dominating this area for their high efficiency. Raúl Mur-Artal et al. \cite{orbslam}\cite{orbslam-pre} proposed a bag of words place recognizer built on DBoW2 with ORB feature \cite{orb-feature} and successful achieved real-time loop closing. In this work, for efficiency and simplicity, we used a similar loop closure detection procedure which is shown in Section \hyperref[back-end-B]{IV-B}.

\section{Unsupervised Monocular Visual Odometry}

In this section, we will introduce our unsupervised approach for monocular visual estimation. 
The training procedure is shown in Fig. 2.

\subsection{Network Architecture}
We name our model NeuralBundler in the sense that, similar to Bundle Adjustment, it performs jointly optimization of both poses and 3D position (depth map) in a neural network fashion. NeuralBundler consists of a pose estimation network and a depth estimation network, which are trained jointly and can be used separately in the testing phase.

The input to the pose estimation network is a stack of views from a sliding window of size $N$, and the output is a windowed pose graph which has $N$ nodes representing the views and $N\times(N-1)$ edges each of which represents the relative 6DoF motion between two views. The network consists of 7 stride-2 convolutions followed by two $1\times1$ convolutions with $6 \times N \times (N-1)$ output channels (corresponding to 3 Euler angles and a 3D translation for each edge in the windowed pose graph). Depth net produces one dense depth map for each RGB image. It has an encoder-decoder shape with skip connections between corresponding encoder and decoder blocks in order to generate high-resolution depth prediction with fine-grained details. The encoder is based on ResNet-50 \cite{resnet} and each decoder block uses a nearest-neighbor upsampling layer followed by a convolutional layer.

\subsection{Loss Function} 

Let's denote $N$ as the total number of views in the input window, $I_i$ as the $i$-th view. $D_i$ denotes the predicted depth map and $T_{ij}$ denotes the predicted relative motion from view $i$ to view $j$. $K$ is the camera's intrinsics. $\mathbb{E}$ is the set of the graph's edges, each of which is represented by a tuple of index: $(i,j)$.  The final loss is a weighted sum of the photometric loss and pose cycle consistency loss.

\subsubsection{Photometric Loss}

Let $p_i$ be the homogeneous coordinates of a pixel in view $i$ and $p_j$ as $p_i$'s projected pixel onto the view $j$. Based on the epipolar geometry, we can obtain $p_j$ from $p_i$ through: 
$$
p_j = KT_{ij}D_{i}K^{-1}p_i 
$$
Then, by applying the differentiable bilinear sampling mechanism proposed in spatial transformer networks \cite{spatial-transformer-networks}, from view $i$ we can synthesis view $j$, which is denoted as $I_{i\rightarrow{j}}$.
Inspired by Godard et al. \cite{monodepth}, the quality of the reconstructed image is measured with the weighted sum of the $l_1$ loss and the single scale structural similarity (SSIM) \cite{ssim} loss. Then the temporal photometric loss is: 
$$
L^{temp}_{pho} = \sum_{(i,j)\in{\mathbb{E}}}  \left[ (1-\alpha) L^{l_1} (I_j,I_{i\rightarrow{j}}) +\\\\
\alpha L^{SSIM}(I_j , I_{i\rightarrow{j}}) \right]  
$$
where $\alpha$ is set to $0.25$. In order to recover metric scale for depth and pose estimation, like \cite{vo-feate}, we apply the same photometric loss between spatial image pairs of the stereo camera.
As shown in Fig. 2, we only use the stereo images from the beginning frame of the window. 
Therefore $I_1$ (default as left image) and $I_{1,R}$ (right image) is the spatial pair. The projection $I_{1,R\rightarrow{1}}$ is synthesized from the known stereo baseline pose and the predicted left depth map. Then the spatial photometric loss is:
$$
L^{spat}_{pho} = (1-\alpha) L^{l_1} (I_1,I_{1,R\rightarrow{1}}) +\\\\
\alpha L^{SSIM}(I_1 , I_{1,R\rightarrow{1}})   
$$

\label{cycle-loss}

\subsubsection{Pose Cycle Consistency Loss}
The windowed pose graph is a complete directed graph containing  $N\times(N-1)$ 6DoF camera motion constraints. Obviously, these constraints may be contradictory to each other, which creates tensions in the pose graph. 
During network training, we relax the windowed pose graph by penalizing a pose cycle consistency loss. Say $(i,j,k)$ are the indexes of three views in the input window. Then the cycle constraint 
$T_{ij}T_{jk}T_{ki}=\mathbf{I}$ holds, where $\mathbf{I}$ is the identity matrix. Let's denote $\mathbb{C}$ as the set of possible cycles in the graph. We penalize the $l_1$ loss: 
$$L_{pos}=\sum_{(i,j,k)\in{\mathbb{C}}} L^{l_1} (  T_{ij}T_{jk}T_{ki}  , \mathbf{I}   )$$
Considering that this loss is similar to the objective of graph optimization (in Section \hyperref[nolinear-opt]{IV-B}), we are somewhat performing a windowed pose graph optimization in a neural network form. Section \hyperref[evaluation]{V-B} demonstrates the efficiency of this loss.

\section{Back-end}

In the back-end, as shown in Fig. 1, we maintain a global pose graph and insert new elements to it each time a new frame is processed by NeuralBundler. We optimize this graph if a loop is established.

\subsection{Pose Graph Construction}
\label{back-end-A}
Pose graph is built on local and loop pose-pose constraints estimated by NeuralBundler. As shown in Fig. 3, local constraints are generated in a sliding window. Loop constraints are obtained in two crossed windows around the loop closure area. Specifically, when there is a loop closure detected between views $I_i$ and $I_j$, the two input windows for NeuralBundler are $<I_i, I_{j-1},...,I_{j-N+1}>$ and $< I_{j}, I_{i-1},...,I_{i-N+1 }>$, where $N$ is the window size.

\begin{figure}[h]
	\centering
	\includegraphics[width=0.43\textwidth]{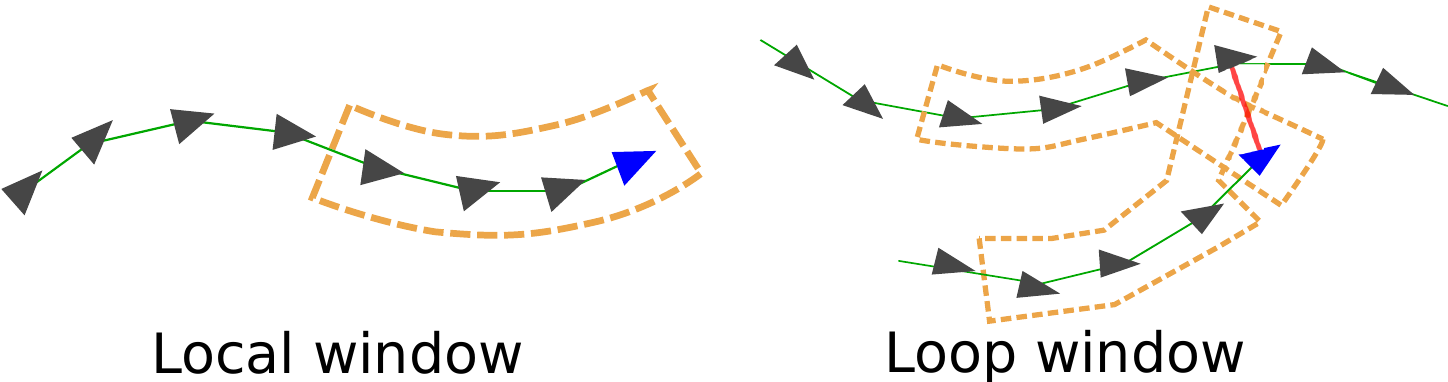}
	\caption{ Image windows for building local and loop pose-pose constraints. The red link indicates a detected loop. Live frames are marked with blue color.
	}
\end{figure}

\label{back-end-B}

The system uses a bags of visual words place recognition tool, based on DBoW2, to perform loop closure detection. We use a predefined  ORB-Vocabulary and ORB-Database which are created off-line with ORB descriptor extracted from a large set of images. Since querying the database with an image will return multiple candidates, similar to \cite{orbslam}, we apply the following procedure to filter the candidates: in order to be accepted, 1) a loop candidate  must be preceded by 6 or more consecutive loop detections, and 2) the loop candidate's corresponding 6DoF transformation must get enough in-liers after several RANSAC iterations.

\subsection{Pose Graph Optimization}
\label{nolinear-opt}

A 3D rigid body transformation  $ \mathrm{\textbf{T} \in SE(3)}$, is defined by:
$$
\mathrm{ \textbf{T}= \left[
	\begin{matrix}
	\textbf{R} & \textbf{t}  \\
	0 & 1  \\
%	7 & 8 & 9
	\end{matrix}
	\right]  \mbox{ with }  \textbf{R} \in{SO(3)} \mbox{ and }  \textbf{t}\in{\mathbb{R}^3} 
}
$$
%Pose Graph Optimization over Sim(3) Constraints: 
%
During optimization, $\mathbf{T}$ is mapped to a minimal representation in $\mathbb{R}^6$ of the associated Lie-algebra through the logarithmic mapping function $\mathrm{\log_{SE(3)}}$ \cite{log-mapping}. Given a pose graph constructed in the proposed way, the error in an edge is defined as:
$$
\mathrm{ \mathbf{e}} _{i,j}=\mathrm{ \log_{SE(3)}}( \mathrm{ \mathbf{T}}^{-1}_{ij} \mathrm{\mathbf{T}}^{-1}_{i}\mathrm{\mathbf{T}}_{j}) 
$$
where $\mathbf{T}_{ij}$ is the relative 6DoF transformation constraints, and the goal is to minimize the total energy:
$$
\chi^2(\mathbf{T}_2,...,\mathbf{T}_m )=\sum_{\mathrm{ \mathbf{T}}_{ij}} (\mathbf{e}^T_{i,j}\mathbf{e}_{i,j})
$$
with respect to the absolute poses $\mathbf{T}_2,...,\mathbf{T}_m $. These absolute poses are initialized through a chain of relative 6DoF transformations starting from the world reference frame $\mathrm{\mathbf{T}}_1$, which is fixed during the optimization. 
We use the Levenberg-Marquardt algorithm implemented in g2o \cite{g2o}  to carry out the optimization.

\section{Experimental Results}

\subsection{Implementation Detail}
We implemented the networks using the publicly available TensorFlow \cite{tensorflow} framework and train it with Tesla P100 GPUs. We trained our model from scratch for 30 epochs, with a mini-batch size of 4 using Adam optimizer \cite{adam}, where $\beta_1=0.9$, $\beta_2=0.999$. We used an initial learning rate of 0.0001 and halve it every 1/5 of the total iterations. The size of the image window for the windowed pose graph estimation network is set to 3 and each image is resized to 416$\times$128. In the testing phase, both the network inference and graph optimization are carried out on an Ubuntu PC equipped with GeForce GTX TITAN X GPU and Intel Core i5 2.4 GHz CPU. The pose estimation network, with about 168k parameters, can be used separately in testing time. Pose estimation network requires less than 400MB GPU memory with over 40 Hz real-time performance.  
For benchmarking, we apply the KIITI odometry datasets \cite{kitti} which contain 11 sequences (00-10) with ground truth trajectory obtained through the IMU/GPS readings. 

\subsection{Trajectory Evaluation}
\label{evaluation}

\begin{figure*} [t]
	\begin{tabular}{ccc}
		
		\includegraphics[width=0.315\textwidth]{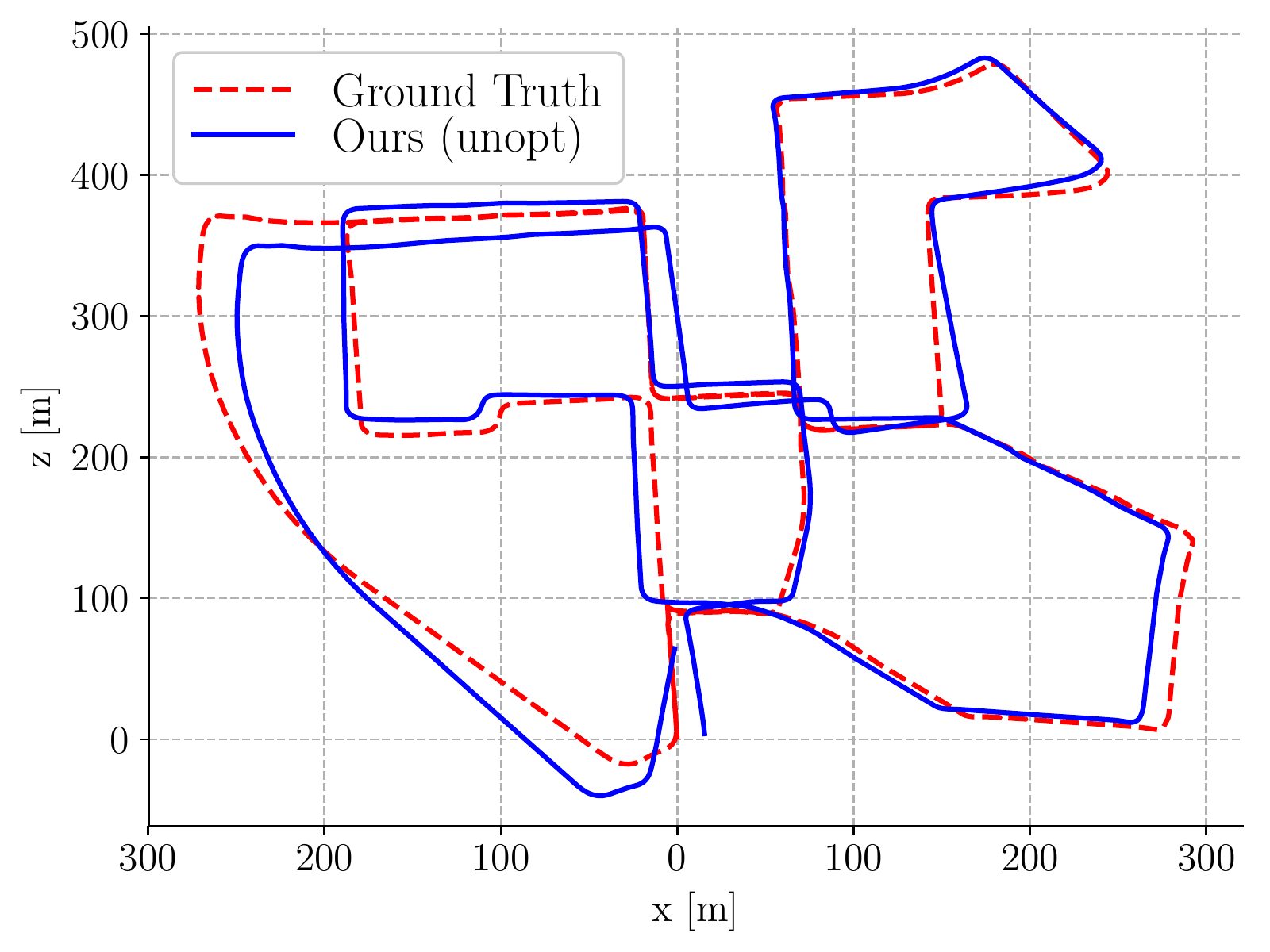} &   \includegraphics[width=0.315\textwidth]{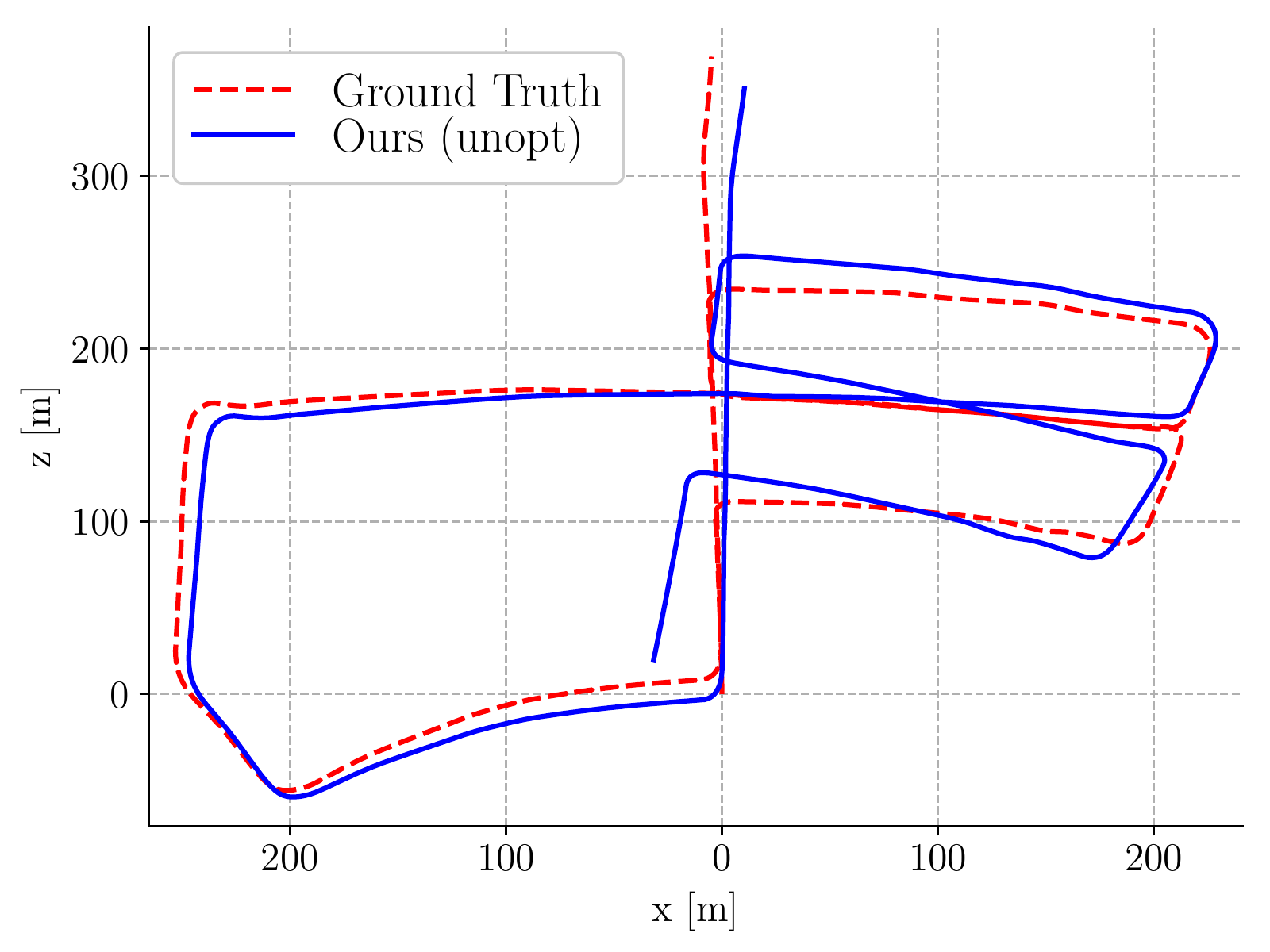} &
		\includegraphics[width=0.315\textwidth]{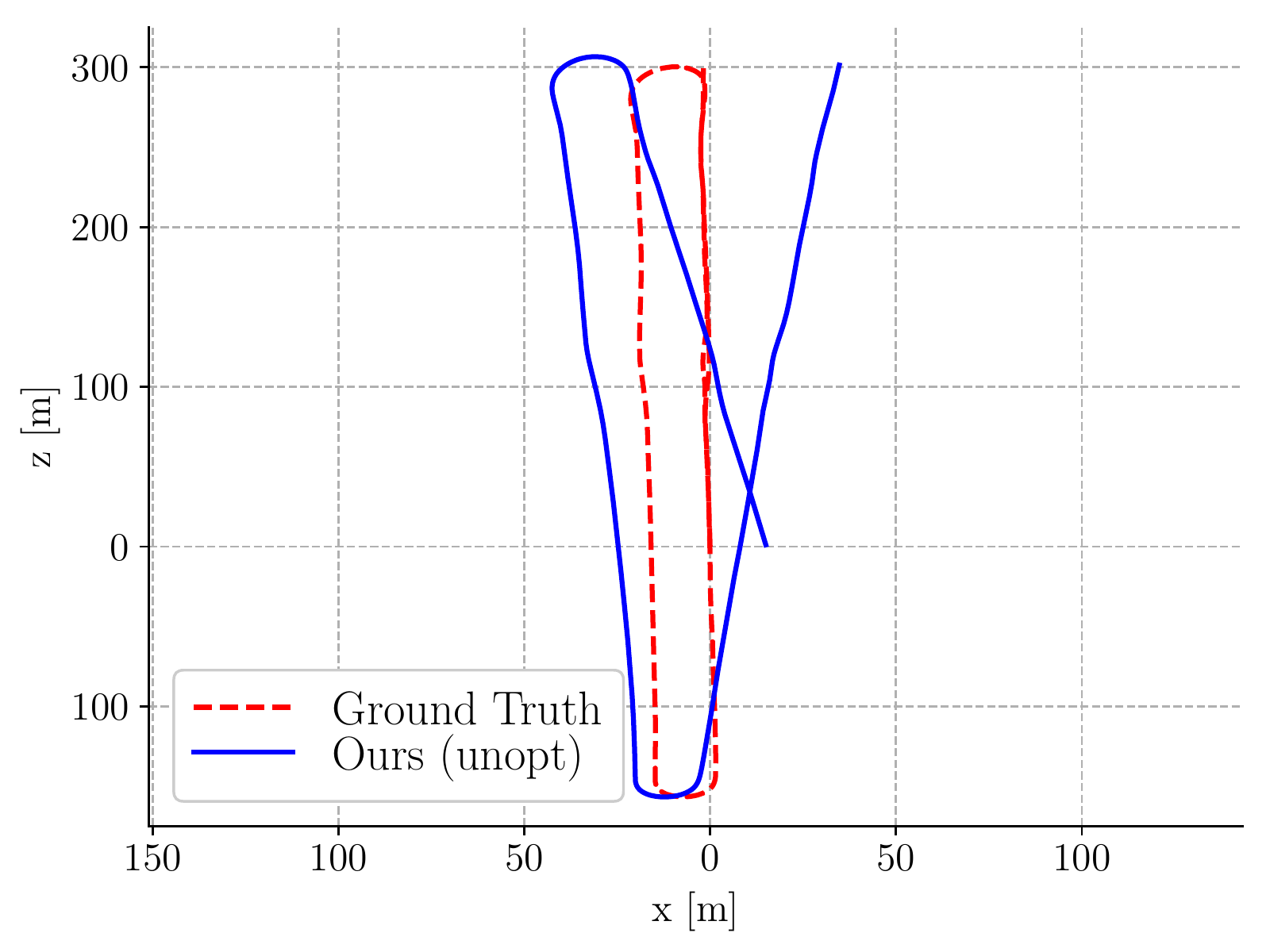}\\
		
		\includegraphics[width=0.315\textwidth]{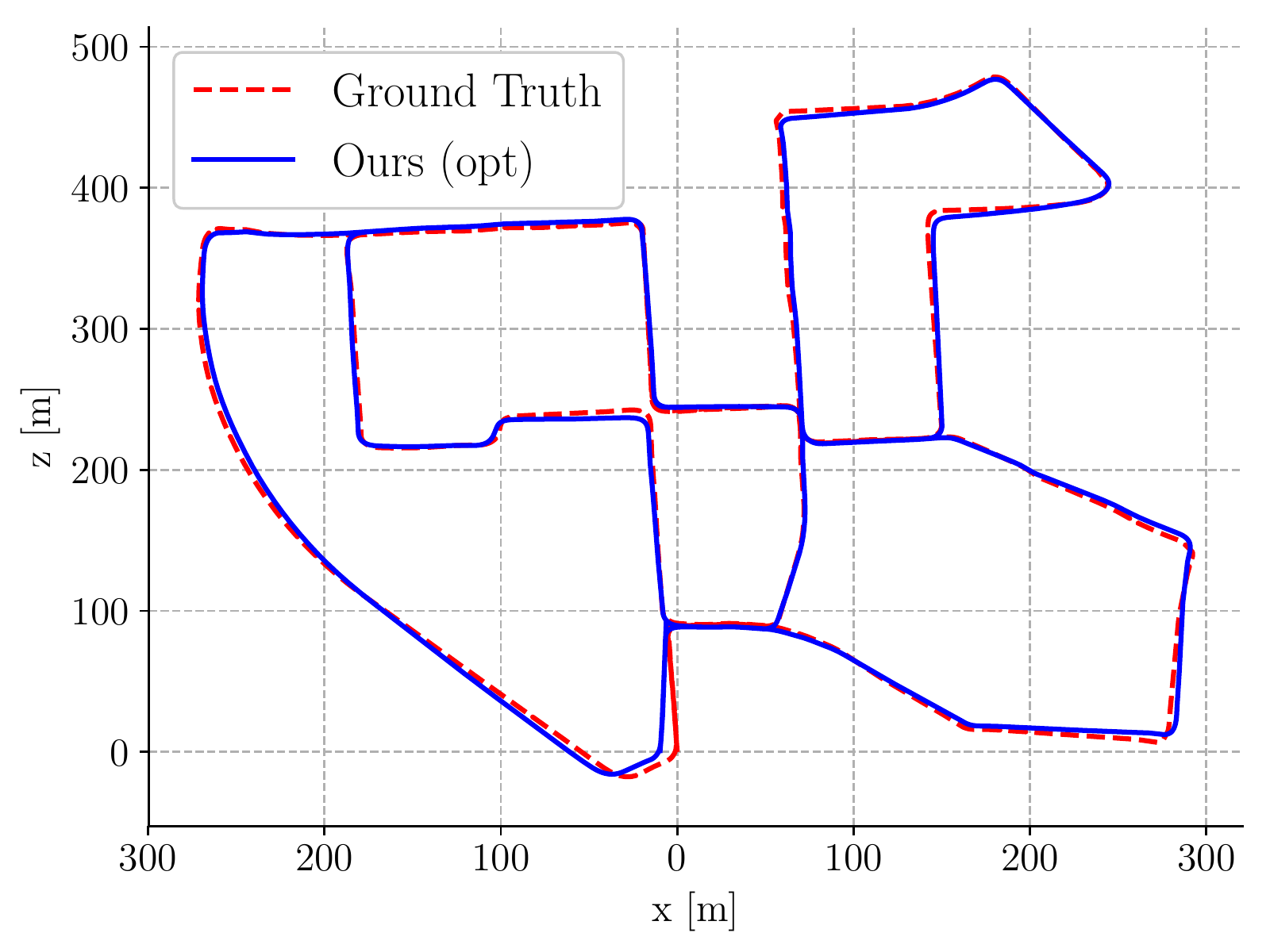} &   \includegraphics[width=0.315\textwidth]{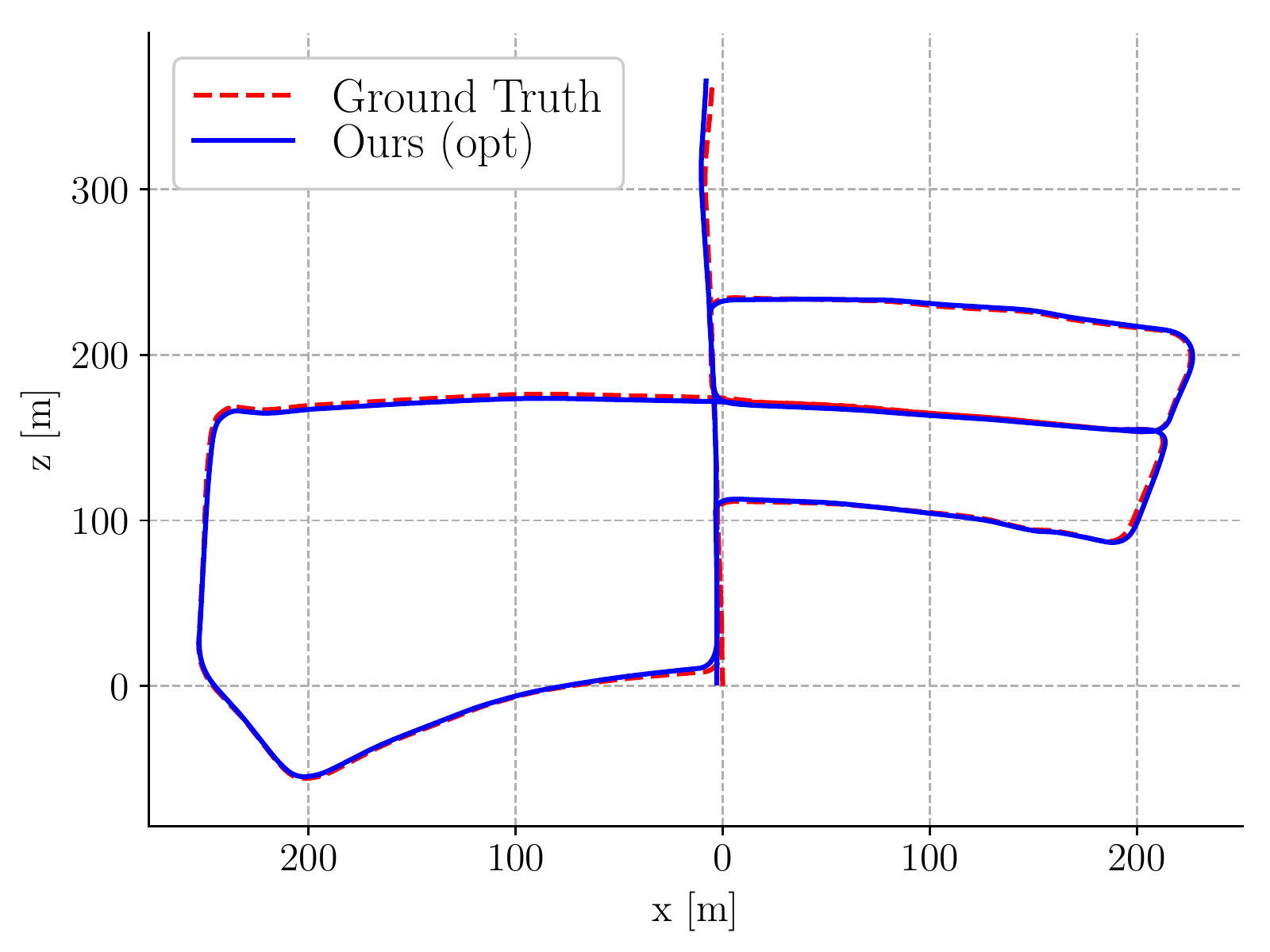} &
		\includegraphics[width=0.315\textwidth]{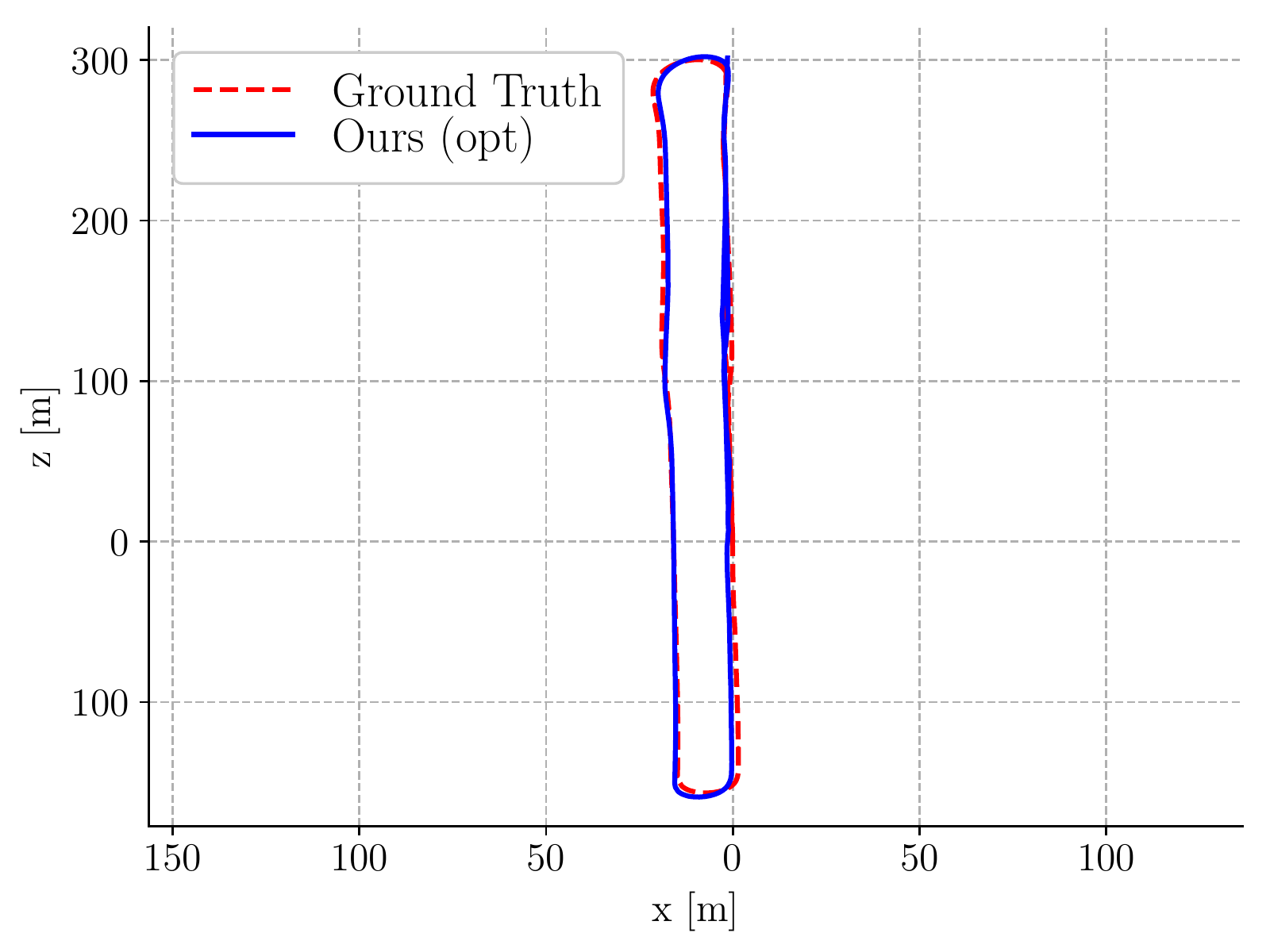}\\
		
	\end{tabular}
	\captionsetup{width=1\linewidth}
	\caption{Results on sequence 00, 05 and 07 from the KITII Odometry dataset. Top row: Raw estimation of NeuralBundler (Trajectory is constructed using the inter-frame motion: $T_{1\rightarrow0}$ from the windowed pose graph, see Section III). Bottom row: After performing loop closing (Graph optimization with loop constraints).}
\end{figure*}

\begin{figure*} [t]
	\begin{tabular}{cccc}
		\includegraphics[width=0.23\textwidth]{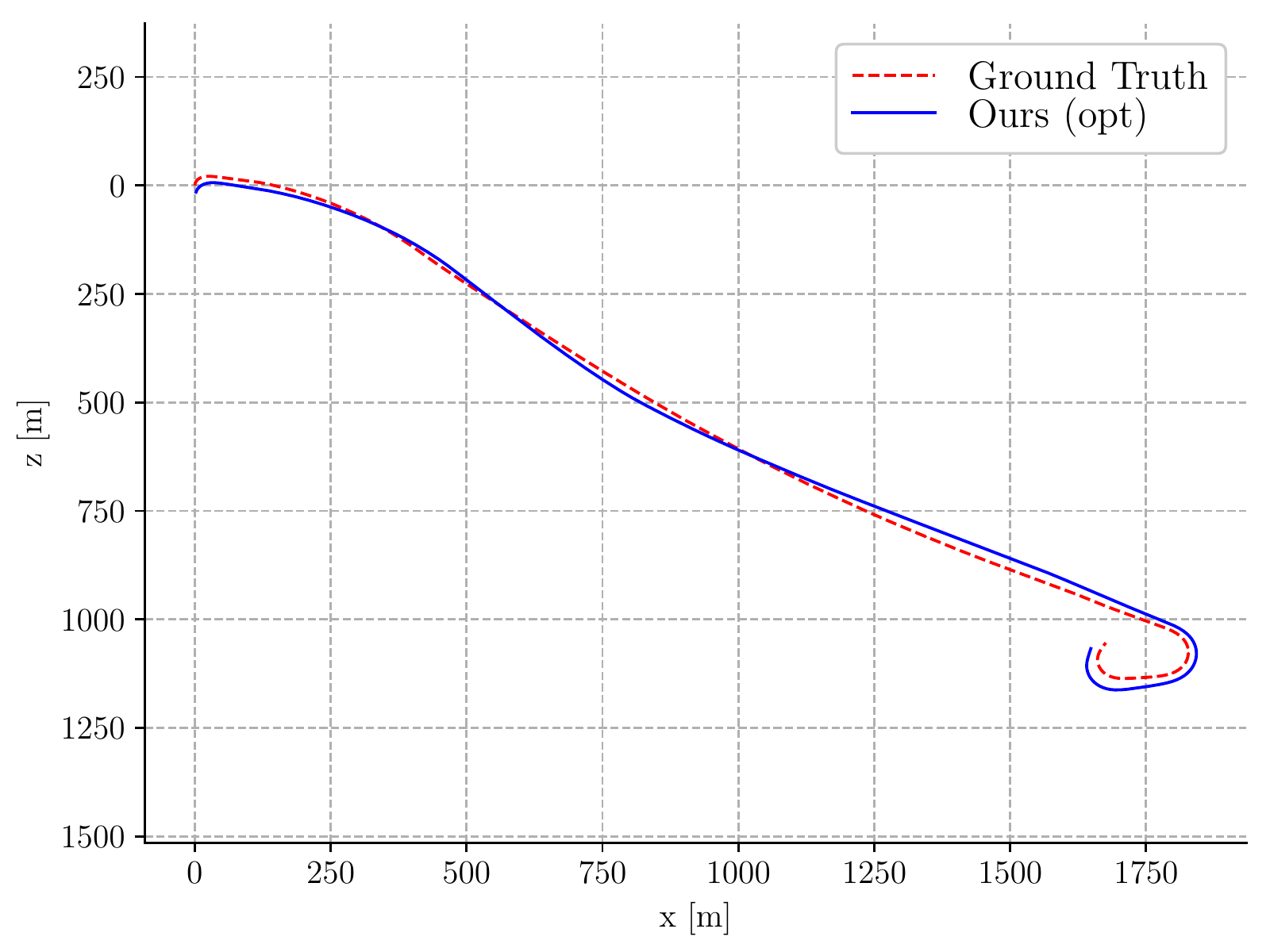} &   
		\includegraphics[width=0.23\textwidth]{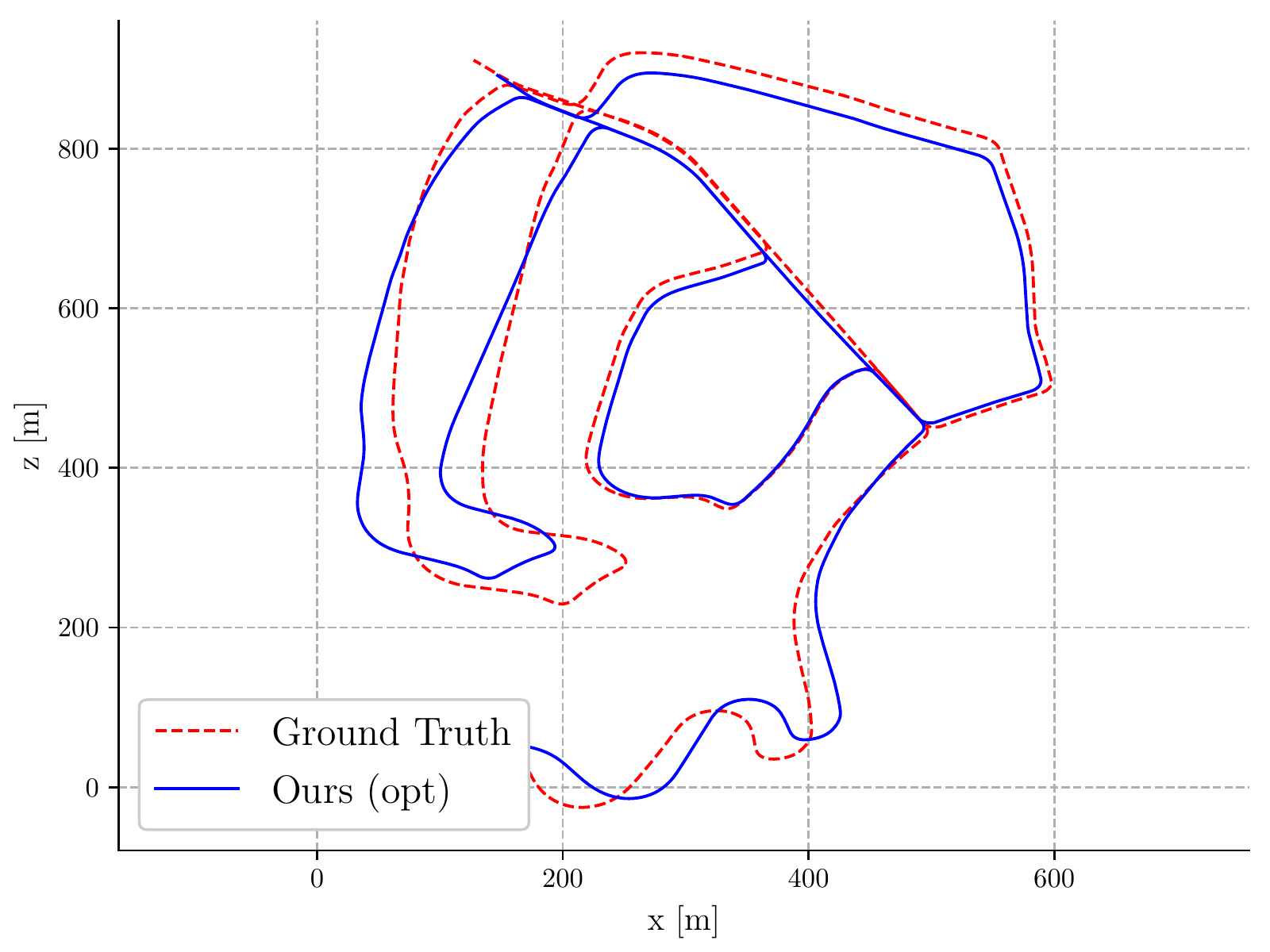} &
		\includegraphics[width=0.23\textwidth]{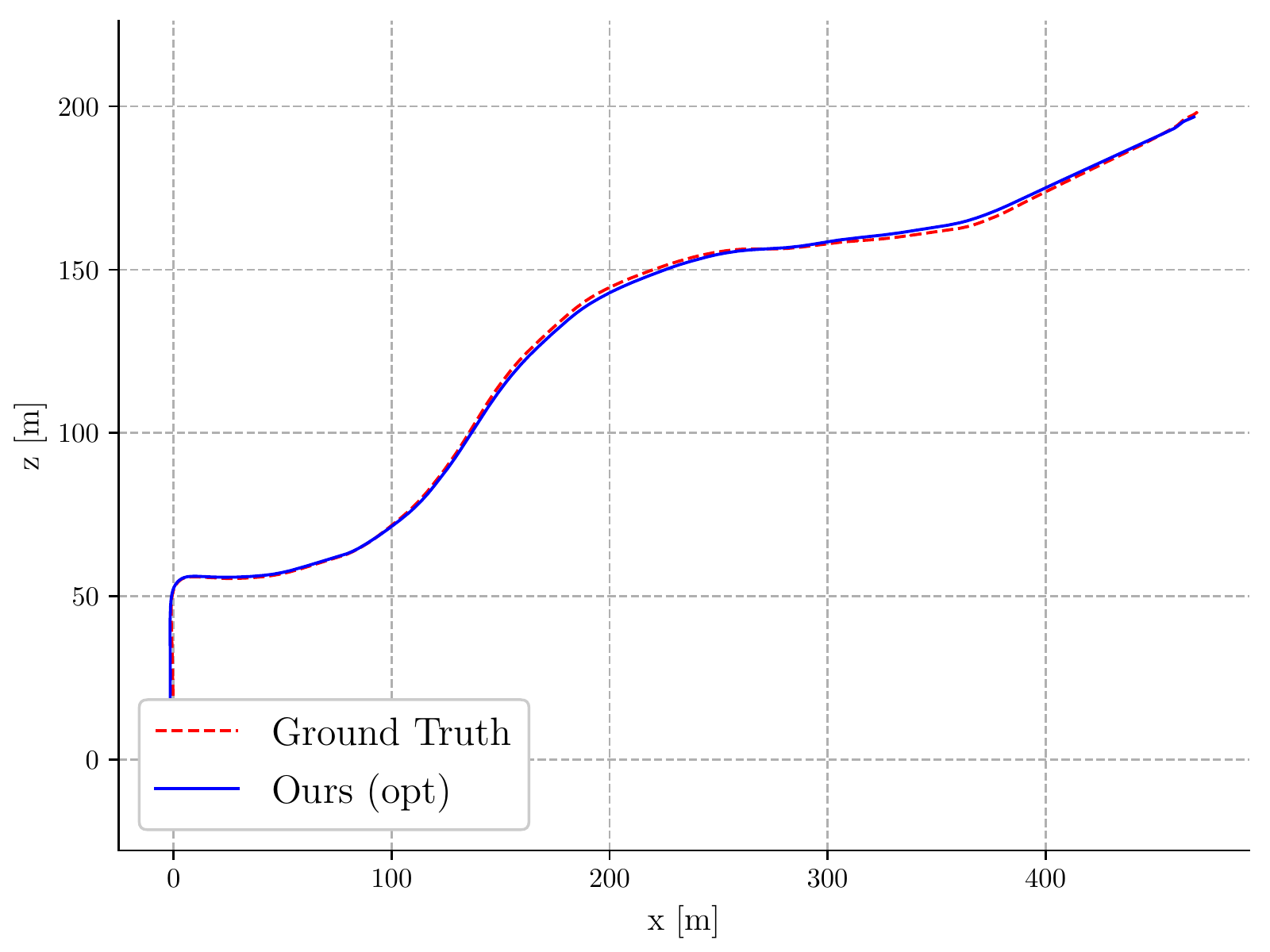} &

		\includegraphics[width=0.23\textwidth]{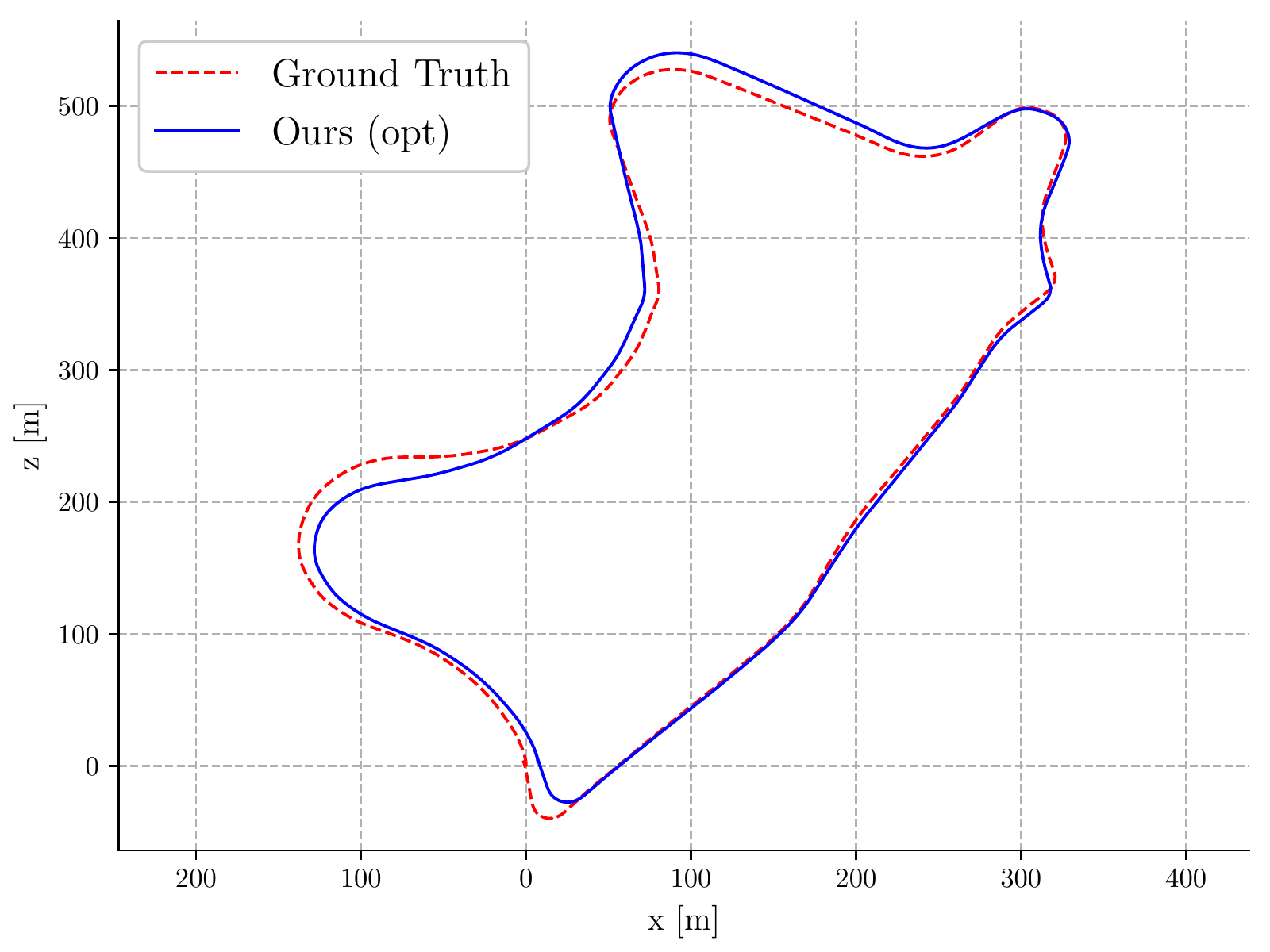} \\
	\end{tabular}
	\captionsetup{width=1\linewidth}
	\caption{ Results of our approach on sequence 01, 02, 03, and 09 from the KITII Odometry dataset.}
	
\end{figure*}

\begin{table*}[t]
	
	%	\captionsetup{width=.85\textwidth}
	
	\caption{ Comparison with unsupervised learning based approaches. $t_{rel}(\%)$ is translational error and $r_{rel}(^o)$ is rotational error. Both are averaged over 100m to 800m intervals. Result of UndeepVO is obtained from \cite{undeepvo} and for SfMLearner \cite{sfmleaner} and Huangying et al. \cite{vo-feate} we ran their pre-trained model. All models share the same training setup. Three variants of our approach is included, NeuralBundler (w/o cycle) : pose cycle consistency loss is not used for training,  NeuralBundler: pose cycle consistency is used for training, NeuralBundler+loop closing: our whole approach. Both the pose cycle consistency loss and loop closing show improvement over the baseline. 
%		Optimization for the sequence without loop closure does not necessarily improve the results. 
%	The highest score for the models (without loop closing) is highlighted.
	}
	\label{table_example}
	\begin{center}
		\renewcommand{\arraystretch}{1.2}
		
		\begin{threeparttable}

			\begin{tabular}{|l |c  |K{0.52cm}  K{0.52cm} K{0.52cm}  K{0.52cm} K{0.52cm}  K{0.52cm}   K{0.52cm}  K{0.52cm}K{0.52cm}  K{0.52cm}   K{0.52cm}  |}
				\hline
				Approach&
			    &00&01&02&03&04&05&06&07&08&09&10 \\

				\hline
				\multirow{2}{5cm}{UnDeepVO \cite{undeepvo}  }   
				& $t_{rel} $ 
				&4.41&--&5.58&--&--&3.4&--&\textbf{3.15}&4.08&--&--
				\\& $r_{rel} $ 
				&1.92&--&2.44&--&--&1.5&--&2.48&1.79&--&--
				\\

				\hline
				\multirow{2}{5cm}{SfMLeaner \cite{sfmleaner}   }   
				& $t_{rel} $ 
				&60.53&35.17&58.75&11.0&4.49&19.14&25.88&21.33&21.90&54.38
				&14.33
			    \\& $r_{rel} $ 
			    &6.12&2.72&3.56&3.94&5.24&4.11&4.81&6.65&2.91&3.21
			    &3.30
				\\
				
				\hline
				\multirow{2}{5cm}{Huangying et al. \cite{vo-feate}  }   
				& $t_{rel} $ 
				&6.62&23.98&6.93&--&3.01&5.12&5.92&7.06&5.85&12.16
				&13.00
				\\& $r_{rel} $ 
				&3.57&1.78&2.39&--&2.02&2.43&2.10&3.8&2.54&3.61
				&3.54
				\\

				\hline
				\multirow{2}{5cm}{Ours, NeuralBundler   (w/o pose cycle loss)}   & $t_{rel} $ 
				&10.31&45.45&7.32&7.97&3.5&5.80&4.64&6.70&5.54&11.98 &\textbf{12.28}
				\\& $r_{rel} $ 
				&2.78&2.13&2.71&4.16&2.01&2.26&3.85&3.46&2.10&3.43&3.97
				\\
				\hline
				\multirow{2}{5cm}{Ours, NeuralBundler    }   & $t_{rel} $ 
				&4.33&\textbf{17.98}&6.89&\textbf{4.51}&\textbf{2.3}&3.91&4.6&3.56&\textbf{4.04}&8.10  &12.9
				\\& $r_{rel} $ 
				&1.85&\textbf{1.44}&2.61&\textbf{2.82}&\textbf{0.87}&1.64&2.85&2.39&\textbf{1.53}&2.81&\textbf{3.17}
				\\
					
				\hline
				\hline
				\hline
%				\multirow{2}{5cm}{Ours, NeuralBundler + opt (w/o loop)}   & $t_{rel} $ 
%				&3.24&--&4.85&--&--&1.83&2.74&3.53&--&6.23&--  
%				\\& $r_{rel} $ 
%				&1.35&--&1.60&--&--&0.7&2.6&2.02&--&2.11&--
%				\\
%				\hline
				\multirow{2}{5cm}{Ours, NeuralBundler + loop closing}   & $t_{rel} $ 
				&\textbf{3.24}&--&\textbf{4.85}&--&--&\textbf{1.83}&\textbf{2.74}&3.53&--&\textbf{6.23}  &--
				\\& $r_{rel} $ 
				&\textbf{1.35}&--&\textbf{1.60}&--&--&\textbf{0.7}&\textbf{2.6}&\textbf{2.02}&--&\textbf{2.11}&--
				\\

				\hline
			\end{tabular}
			%			\begin{tablenotes}
			%				\item[*] with loop closure.
			%				\item[X] unavailable due to tracking failure.
			%			\end{tablenotes}
		\end{threeparttable}
	\end{center}
\end{table*}

\begin{table*}[t]
	
	%	\captionsetup{width=.85\textwidth}
	
	\caption{RMSE error of estimated trajectories on KITTI Odometry Dataset. All the methods listed in the table perform loop closing if any loop is detected.  
	Results of ORB-SLAM and ORB-SLAM + Global BA (20 its.) are taken from \cite{orbslam}. 
	ORB-SLAM: perform optimization on the Essential Graph \cite{orbslam}, ORB-SLAM + Global BA (20 its): perform 20 iterations of global Bundler Adjustment afterward.
	For trajectory alignment, we use 6DoF transformations and ORB-SLAM uses 7DoF transformations. Transformations are optimized to achieve the best alignment.
	}
	\label{table_example}
	\begin{center}
		\renewcommand{\arraystretch}{1.5}
		
		\begin{threeparttable}
			
			\begin{tabular}{|l| K{0.5cm}  K{0.5cm} K{0.5cm}  K{0.5cm} K{0.5cm}  K{0.5cm}  K{0.5cm}  K{0.5cm}K{0.5cm}  K{0.5cm} K{0.5cm} | }
				\hline
				&00$^*$&02$^*$&05$^*$&06$^*$&07$^*$&09$^*$&01&03&04&08&10 \\ 
				
				\cline{2-12}
				Approach& \multicolumn{11}{c|}{RMSE (m)} \\
				\hline

				Ours, NeuralBundler + Opt &\textbf{4.36}&52.75&\textbf{2.57}&\textbf{3.71}&\textbf{1.67}&14.85&\textbf{57.4}&\textbf{1.43}&\textbf{0.82}&\textbf{16.4}&11.2\\
				
				\hline

				ORB-SLAM  			  			&6.68&21.75&8.23&14.68&3.36&7.62&X&1.59&1.79&46.58&\textbf{8.68}\\
				ORB-SLAM + Global BA (20 its.) 			&5.33&\textbf{21.28}&4.85&12.34&2.26&\textbf{6.62}&X&1.51&1.62&46.68&8.80\\	
				\hline
				
			\end{tabular}
			\begin{tablenotes}
				\item[*] with loop closure.
				\item[X] unavailable due to tracking failure.
				
			\end{tablenotes}
		\end{threeparttable}
	\end{center}
\end{table*}
Qualitative comparisons of our trajectories and the ground truth are shown in Fig. 4, Fig. 5 and Fig. 6. 
We align the trajectories with the ground truth through rigid body transformations on ${\rm SE(3)}$ (No scaling is applied for that our VO system can recover the metric scale). 
Sequences 00, 02, 05, 06, 07, 09 contain loops that were correctly detected and closed by our system. Fig. 4 shows the effectiveness of our loop closing approach on sequence 00, 05 and 06. The top row shows the raw VO estimation of NeuralBundler, and the bottom row shows corresponding results after performing loop closing. The raw VO estimation gives decent local accuracy while with gradually accumulated drift. After performing graph optimization with correctly detected loops, the drift is significantly reduced, leading to good fits with the ground truth. 

For evaluation, first, we use the error metrics proposed in \cite{kitti} to compare our approach with other unsupervised learning based visual odometry systems: UnDeepVO \cite{undeepvo}, SfMLeaner \cite{sfmleaner} and Huangying et al. \cite{vo-feate}. As shown in table I. For ablation study, our NeuralBundler consistently yields better results than the variant without pose cycle consistency loss. NeuralBundler outperforms SfMLeaner and Huangying et al. with a large margin and shows comparable results to UnDeepVO (which trains the network with extra supervisions from point cloud alignment and stereo camera's left-right pose consistency \cite{undeepvo}). Applying loop closing significantly reduced the error, while graph optimization without loop constraints (not shown) does not necessarily lead to better results, which makes clear the need of loop closures for global drift correction.

 \begin{figure} [b]
 	\begin{tabular}{cc}
 		\includegraphics[width=0.22\textwidth]{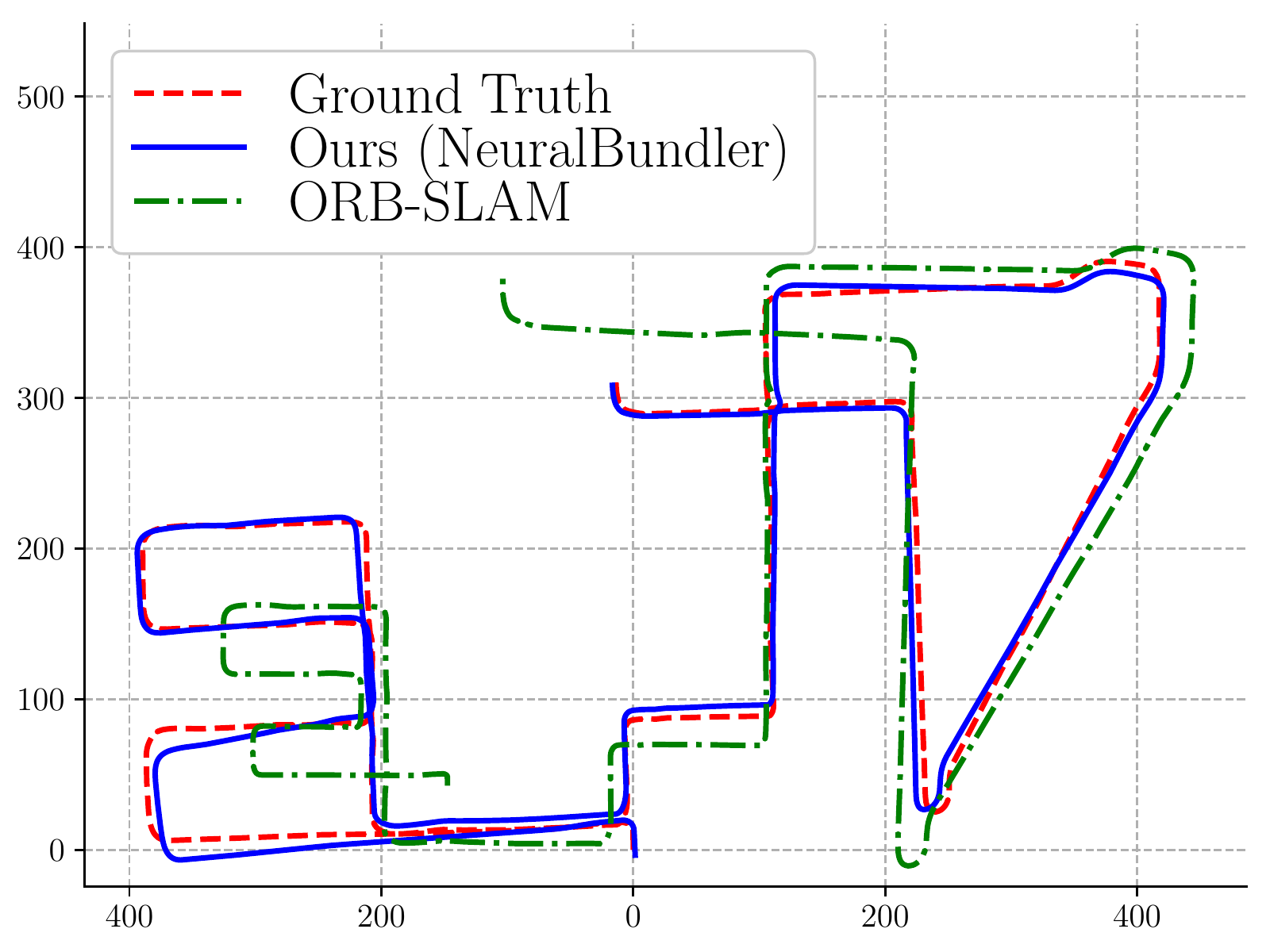} &   
 		\includegraphics[width=0.22\textwidth]{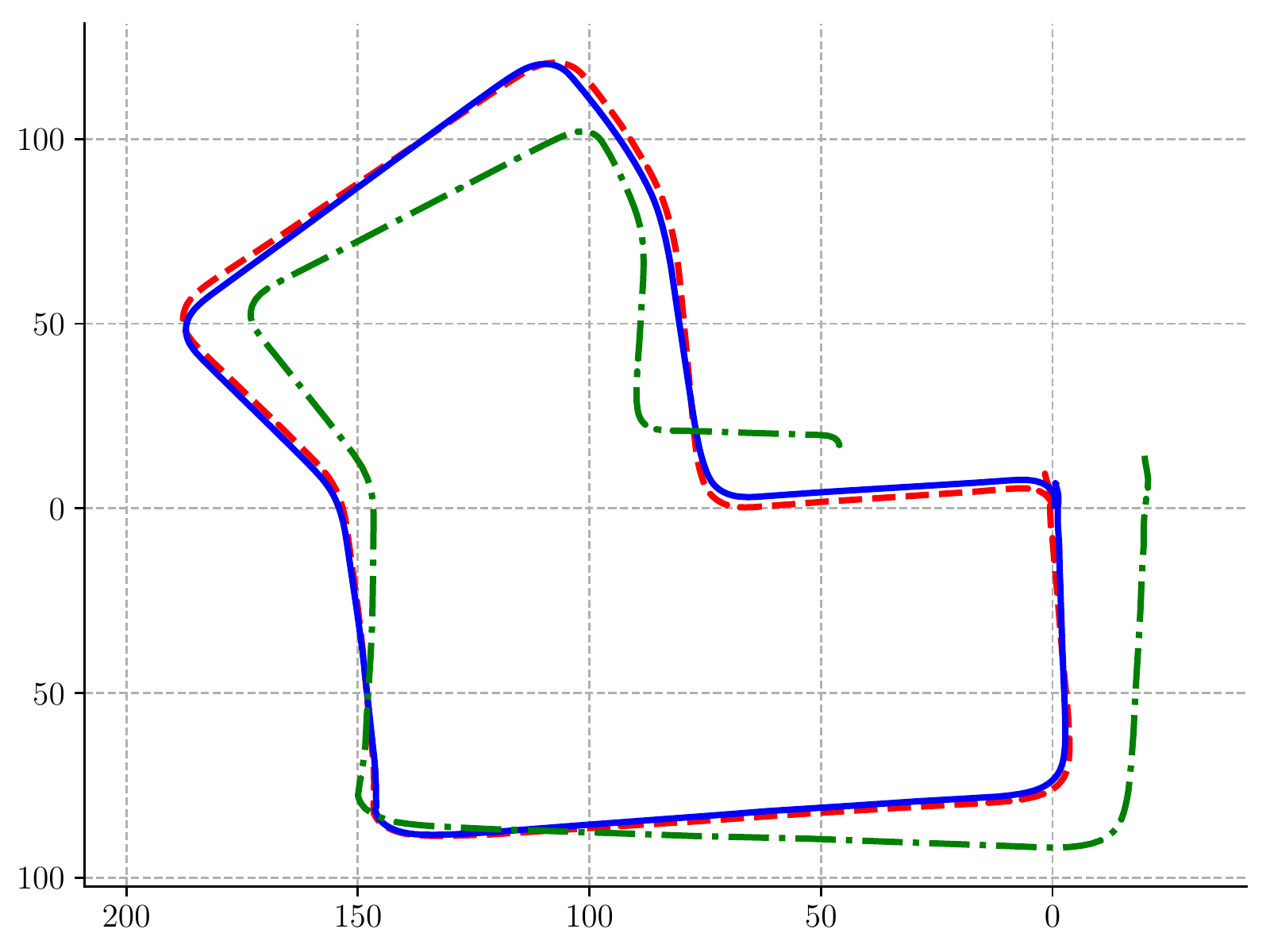} \\
 	\end{tabular}
 	\captionsetup{width=1\linewidth}
 	\caption{ Results on sequence 08 (left) and 07 (right). Sequence 08 does not contain a loop. For sequence 07, we turn off the loop closing thread of ORB-SLAM. ORB-SLAM suffers from severe scale drift, and heavily rely on loop closing to eliminate scale drift. }
 	
 \end{figure}

Then, we compare the results of our method with the state of the art feature-based ORB-SLAM. This time, we directly assess the overall translational RMSE error of the trajectory. As shown in table II, our method achieves smaller overall translational error for most of the sequences (8 out of 11). This makes sense considering that though both methods use monocular image, our approach is more robust to scale drift. Fig. 6 shows the estimated trajectory on sequence 07 and 08. Geometry based ORB-SLAM suffers from sever scale drift, while NeuralBundler achieves superior performance on eliminating the scale drift. 
ORB-SLAM encounters tracking failure on sequence 01 (a high way sequence). 
However, partially due to the fact that geometric methods know exactly the pixel-pixel correspondence, ORB-SLAM yields smaller rotational errors, which could explain our 3 lost cases.

\section{Conclusion and Discussion}

In this work, we have presented a new monocular VO system which has an unsupervised learning based front-end called NeuralBundler and a graph optimization back-end. 
Results on KITTI odometry have proved the effectiveness of pose cycle consistency loss. Our whole approach can achieve efficient loop closure and show better overall RMSE than ORB-SLAM for most sequences in KITTI odometry datasets. A number of major challenges are yet to be addressed:
\begin{itemize}

\item We have not applied any key-frame selection and edge removal procedure yet. In our method, every frame becomes a "key-frame", and each node is connected with 12 edges (in the case that window size $N$ is 3 and the node is not on the head or tail), which lead to a much denser and more complex pose graph than the SLAM's version. Though real-time loop closing is achieved, such techniques are still necessary for the sake of scalability and efficiency.
\item The model is trained on sequences with fixed camera intrinsics, fixed input image size, and from limited scenes (car driving on the road). We need to solve the domain adaptation problem when applying to different situations.
%\item We still need to extract and maintain features in the background in order to use DBoW2 tool for loop closure detection, which considerably consumes memory and CPU power. The solution is to directly perform scene recognition with the deep neural network. We put this to future work.
\item Unsupervised VO still lose to traditional methods on Mean Rotation/Translation accuracy. A possible solution is to use the coarse-to-fine strategy to optimize the VO estimation as in \cite{DeepTAM}.
\end{itemize}

\section{Acknowledgments}

YL would like to thank Tinghui Zhou and Huangying Zhan for helpful discussions.

\end{document}